\documentclass[review]{elsarticle}

\usepackage{lineno,hyperref}
\usepackage{times}
\usepackage{url}
\usepackage{latexsym}
\usepackage{amssymb}
\usepackage{amsmath}
\DeclareMathOperator*{\argmax}{argmax} 

\usepackage{cleveref}

\usepackage{stmaryrd}
\usepackage{graphicx}
\usepackage{adjustbox}
\usepackage{listings}
\usepackage{caption,setspace}
\usepackage{diagbox} 
\usepackage{booktabs} 
\usepackage[dvipsnames]{xcolor} 
\usepackage{rotating}
\usepackage{array,multirow}
\usepackage{hhline}
\usepackage{todonotes}
\usepackage{flushend}
\usepackage[inline]{enumitem}
\usepackage{csquotes}

\usepackage{color}

\usepackage{mdframed}
\newmdenv[
  skipabove=\topsep,
  skipbelow=\topsep
]{reviewercomment}

\usepackage[normalem]{ulem} 

\usepackage{changepage}  

\lstdefinestyle{base}{
  moredelim=**[is][\color{red}]{@}{@} , 
  moredelim=**[is][\color{OliveGreen}]{!}{!},
    moredelim=**[is][\color{purple}]{)}{)},
  moredelim=**[is][\color{blue}]{?}{?},
   moredelim=**[is][\color{orange}]{(}{(},
  moredelim=[is][\bfseries]{[*}{*]}
}

\newcommand{\eg}{e.g., }
\newcommand{\ie}{i.e., }
\newcommand{\figref}[1]{Fig.~\ref{#1}}    
\newcommand{\tabref}[1]{Table~\ref{#1}}
\newcommand{\Tabref}[1]{Table~\ref{#1}}
\newcommand{\secref}[1]{Section~\ref{#1}}
\modulolinenumbers[5]

\newcommand{\cev}[1]{\reflectbox{\ensuremath{\vec{\reflectbox{\ensuremath{#1}}}}}}

\journal{Expert Systems with Applications}

\biboptions{authoryear}

\begin{document}
\setlength{\abovedisplayskip}{3pt}
\setlength{\belowdisplayskip}{3pt}

\begin{frontmatter}

\title{An attentive neural architecture for joint segmentation and parsing and
its application to real estate ads}

\author[]{Giannis Bekoulis\corref{cor1}}
\ead{giannis.bekoulis@ugent.be}
\author[]{Johannes Deleu}
\ead{johannes.deleu@ugent.be}
\author[]{Thomas Demeester}
\ead{ thomas.demeester@ugent.be}
\author[]{Chris Develder}
\ead{chris.develder@ugent.be}
\cortext[cor1]{Corresponding author}
\address{Ghent University -- imec, IDLab, Department of Information Technology,\\
Technologiepark Zwijnaarde 15, 9052 Ghent, Belgium}

\begin{abstract}

\noindent In processing human produced text using natural language processing (NLP) techniques, two fundamental subtasks that arise are 
\begin{enumerate*}[label=(\roman*)]
\item \emph{segmentation} of the plain text into meaningful subunits (\eg entities), and
\item \emph{dependency parsing}, to establish relations between subunits.
\end{enumerate*}
Such structural interpretation of text provides essential building blocks for upstream expert system tasks: \eg from interpreting textual real estate ads, one may want to provide an accurate price estimate and/or provide selection filters for end users looking for a particular property --- which all could rely on knowing the types and number of rooms, etc.
In this paper we develop a relatively simple and effective neural joint model that performs both segmentation and dependency parsing together, instead of one after the other as in most state-of-the-art works.
We will focus in particular on the real estate ad setting, aiming to convert an ad to a
structured description, which we name \emph{property tree}, comprising the tasks of
\begin{enumerate*}[label=(\arabic*)]
\item identifying important entities of a property (\eg rooms) from classifieds and
\item structuring them into a 
tree format. 
\end{enumerate*}
In this work, we propose a new joint model that is able to tackle the two tasks simultaneously and construct the \emph{property tree} by
\begin{enumerate*}[label=(\roman*)]
\item avoiding the error propagation that would arise from the subtasks one after the other in a pipelined fashion, and
\item exploiting the interactions between the subtasks.
\end{enumerate*}
For this purpose, we perform an extensive comparative study of the pipeline methods and the new proposed joint model, reporting 
an improvement of over three percentage points in the overall edge $F_1$ score of the property tree. 
Also, we propose attention methods, to encourage our model to focus on salient tokens during the construction of the \emph{property tree}. Thus we experimentally demonstrate the usefulness of attentive neural architectures for the proposed joint model, showcasing 
a further improvement of two percentage points
in edge $F_1$ score for our application. 
While the results demonstrated are for the particular real estate setting, the model is generic in nature, and thus could be equally applied to other expert system scenarios requiring the general tasks of both
\begin{enumerate*}[label=(\roman*)]
\item detecting entities (\emph{segmentation}) and
\item establishing relations among them (\emph{dependency parsing}).
\end{enumerate*}
\end{abstract}


\end{frontmatter}


\section{Introduction}
\label{sec:intro}

\noindent Many consumer-oriented digital applications rely on
input data provided by their target audience. For instance, real estate websites gather property descriptions
for the offered classifieds, either from realtors
or from individual sellers. In such cases, it is
hard to strike the right balance between structured
and unstructured information: enforcing restrictions
or structure upon the data format (\ie predefined form) may reduce the
amount or diversity of the data, while unstructured data (\ie raw text) may require
non-trivial (\ie hard to automate) transformation
to a more structured form to be useful/practical
for the intended application. In the real estate domain, textual advertisements are an extremely useful but highly unstructured way of representing real estate properties.
However, structured descriptions of the advertisements are very helpful, 
\eg for real estate agencies to suggest the most appropriate sales/rentals for their customers, while keeping human reading effort limited. For example, special search filters, which are usually used by clients, cannot be directly applied to textual advertisements.
On the contrary, a structured representation of the property (\eg a tree format of the property) enables the simplification of the unstructured textual information by applying specific filters (\eg based on
the number of bedrooms, number of floors, or the requirement
of having a bathroom with a toilet on the
first floor), and it also benefits other related applications
such as automated price prediction~\citep{pace:00,nagaraja:11}.

The new real estate structured prediction problem as defined by~\cite{bekoulis:17} has as main goal to construct the tree-like representation of the property (\ie the \emph{property tree}) based on its natural language description. This can be approached as a relation extraction task by a pipeline of separate subtasks, comprising
\begin{enumerate*}[label=(\roman*)]
\item named entity recognition (NER)~\citep{nadeau:07} and
\item relation extraction~\citep{bach:07}.
\end{enumerate*}
Unlike previous studies~\citep{li:14, miwa:16} on relation extraction, in the work of~\cite{bekoulis:17}, the relation extraction module is replaced by dependency parsing.
Indeed, the relations that together define the structure of the house should form a tree, where entities are \emph{part-of} one another (\eg a floor is \emph{part-of} a house, a room is \emph{part-of} a floor). This \emph{property tree} is structurally similar to a parse tree.
Although the work of~\cite{bekoulis:17} is a step towards the construction of the \emph{property tree}, it follows a pipeline setting, which suffers from two serious problems:
\begin{enumerate*}[label=(\roman*)]
\item error propagation between the subtasks, \ie NER and dependency parsing, and
\item cross-task dependencies are not taken into account, \eg terms indicating relations (includes, contains, etc.)\ between entities that can help the NER module are neglected.
\end{enumerate*}
Due to the unidirectional nature of stacking the two modules (\ie NER and dependency parsing) in the pipeline model, there is no information flowing from the dependency parsing to the NER subtask. This way, the parser is not able to influence the predictions of the NER.
Other studies on similar tasks~\citep{li:14,kate:10} have considered the two subtasks jointly. They simultaneously extract entity mentions and relations between them usually by implementing a beam-search on top of the first module (\ie NER), but these methods require the manual extraction of hand-crafted features.
Recently, deep learning with neural networks has received much attention and several approaches \citep{miwa:16,zheng:17} apply long short-term memory (LSTM) recurrent neural networks and convolutional neural networks (CNNs) to achieve state-of-the-art performance on similar problems. Those models rely on shared parameters between the NER and relation extraction components, whereby the NER module is typically pre-trained separately, to improve the training effectiveness of the joint model. 

In this work, we propose a new joint model to solve the real estate structured prediction problem. Our model is able to learn the structured prediction task without complicated feature engineering. Whereas previous studies \citep{miwa:16,zheng:17,li:16,li:17} on joint methods focus on the relation extraction problem, we construct the \emph{property tree} which comes down to solving a dependency parsing problem, which is more constrained and hence more difficult.
Therefore, previous methods are not directly comparable to our model and cannot be applied to our real estate task out-of-the-box. In this work, we treat the two subtasks as one by reformulating them into a head selection problem~\citep{zhang:16}.

This paper is a follow-up work of \cite{bekoulis:17}. Compared to the conference paper that introduced the real estate extraction task and applied some basic state-of-the-art techniques as a first baseline solution, we now introduce: 
\begin{enumerate*}[label=(\roman*)]
\item advanced neural models that consider the two subtasks jointly and
\item modifications to the dataset annotation representations as detailed below.
\end{enumerate*}
More specifically, the main contributions of this work are the following:
\begin{itemize}
\item We propose a new joint model that encodes the two tasks of identifying entities as well as dependencies between them, as a single head selection problem, without the need of parameter sharing or pre-training of the first entity recognition module separately. Moreover, instead of 
\begin{enumerate*}[label=(\roman*)]
\item predicting unlabeled dependencies and 
\item training an additional classifier to predict labels for the identified heads~\citep{zhang:16},
\end{enumerate*} 
our model already incorporates the dependency label predictions in its scoring formula.
\item We compare the proposed joint model against established pipeline approaches and report an $F_1$ improvement of 1.4\% in the NER and 6.2\% in the dependency parsing subtask, 
corresponding to an overall edge $F_1$ improvement of 3.4\% in the property tree.

\item Compared to our original dataset~\citep{bekoulis:17}, we introduce two extensions to the data:
\begin{enumerate*}[label=(\roman*)]
\item we consistently assign the first mention of a particular entity in order of appearance in the advertisement as the main mention of the entity. This results in an $F_1$ score increase of about 3\% and 4\% for the joint and pipeline models, respectively. 
\item We add the \emph{equivalent} relation to our annotated dataset to explicitly express that several mentions across the ad may refer to the same entity. 
\end{enumerate*}
\item We perform extensive analysis of several attention mechanisms that enable our LSTM-based model to focus on informative words and phrases, reporting an improved $F_1$ performance of about 2.1\%.
\end{itemize}

The rest of the paper is structured as follows. In \secref{sec:related}, we review the related work. \secref{problem_definition} defines the problem and in \secref{sec:methodology}, we describe the methodology followed throughout the paper and the proposed joint model. The experimental results are reported in \secref{sec:results_discussion}. Finally, \secref{sec:conclusions} concludes our work.

\section{Related work}
\label{sec:related}

\noindent The real estate structured prediction problem from textual advertisements can be broken down into the sub-problems of
\begin{enumerate*}[label=(\roman*)]
\item sequence labeling (identifying the core parts of the property) and
\item non-projective dependency parsing (connecting the identified parts into a tree-like structure)~\citep{bekoulis:17}.
\end{enumerate*}
One can address these two steps either one by one in a pipelined approach, or simultaneously in a joint model.
The pipeline approach is the most commonly used approach \citep{bekoulis:17,fundel:07,gurulingappa:12}, treating
the two steps independently and propagating
the output of the sequence labeling subtask (\eg named entity recognition) \citep{chiu:15,lample:16} to the relation classification module \citep{santos:15,xu:15}.
Joint models are able to simultaneously extract entity mentions and relations between them \citep{li:14,miwa:16}.
In this work, we propose a new joint model that is able to recover the tree-like structure of the property and 
frame it as a dependency parsing problem,
given the non-projective tree structure we aim to output.
We now present related works for the sequence labeling and dependency parsing subtasks, as well as for the joint models.

\subsection{Sequence labeling}
\label{sec:sequence_labeling}
\noindent Structured prediction problems become challenging due to the large output space. 
Specifically in NLP, sequence labeling (\eg NER) is the task of identifying the entity mention boundaries and assigning a categorical label (\eg POS tags) for each identified entity in the sentence.
A number of different methods have been proposed, namely Hidden Markov Models (HMMs)~\citep{rabiner:86}, Conditional Random Fields (CRFs)~\citep{crf:01}, Maximum Margin Markov Network (M$^3$N) \citep{taskar:04}, generalized support vector machines for structured output (SVM$^{struct}$) \citep{tsochantaridis:04} and Search-based Structured Prediction (SEARN) \citep{daume:09}. Those methods heavily rely on hand-crafted features and an in-depth review can be found in \cite{nguyen:07}.
Several variations of these models that also require manual feature engineering have been used in different application settings (\eg biology, social media context) and languages (\eg Turkish) \citep{jung:12,kucuk:12,atkinson:12,konkol:2015}.
Recently, deep learning with neural networks has been succesfully applied to NER.~\cite{collobert:11} proposed to use a convolutional neural network (CNN) followed by a CRF layer over a sequence of word embeddings. 
Recurrent Neural Networks (RNNs) constitute another neural network architecture that has attracted attention, due to the state-of-the-art performance in a series of NLP tasks (\eg sequence-to-sequence~\citep{sutskever:14}, parsing~\citep{kiperwasser:16}). In this context,~\cite{gillick:15} use a sequence-to-sequence approach for modeling the sequence labeling task. In addition, several variants of combinations between LSTM and CRF models have been proposed \citep{lample:16,huang:15,ma:16} achieving state-of-the-art performance on publicly available datasets.

\subsection{Dependency parsing}
\noindent Dependency parsing is a well studied task in the NLP community, which aims to analyze the grammatical structure of a sentence. We approach the problem of the \emph{property tree} construction as a dependency parsing task \ie to learn the dependency arcs of the classified. There are two well-established ways to address the dependency parsing problem, via graph-based and transition-based parsers.
\\
\textbf{Graph-based:} In the work of~\cite{mcdonald:05,mcdonald:06} dependency parsing requires the search of the highest scoring maximum spanning tree in graphs for both projective (dependencies are not allowed to cross) and non-projective (crossing dependencies are allowed) trees with the Eisner algorithm~\citep{eisner:96} and the Chu-Liu-Edmonds algorithm~\citep{chu:65,edmond:68} respectively.  It was shown that exploiting higher-order information (\eg siblings, grand-parental relation) in the graph, instead of just using first-order information (\ie parent relations)~\citep{carreras:07, zhang:12} may yield significant improvements of the parsing accuracy but comes at the cost of an increased model complexity.~\cite{koo:07} made an important step towards globally normalized models with hand-crafted features, by adapting the Matrix-Tree Theorem (MTT)~\citep{tutte:01} to train over all non-projective dependency trees. We explore an MTT approach as one of the pipeline baselines. Similar to recent advances in neural graph-based parsing~\citep{zhang:16,kiperwasser:16,wang:16}, we use LSTMs to capture richer contextual information compared to hand-crafted feature based methods. Our work is conceptually related to~\cite{zhang:16}, who formulated the dependency parsing problem as a head selection problem. We go a step further in that direction, in formulating the joint parsing and labeling problem in terms of selecting the most likely combination of head and label.
\\
\textbf{Transition-based:} Transition-based parsers~\citep{yamada:03,nivre:06} replace the exact inference of the graph-based parsers by an approximate but faster inference method. The dependency parsing problem is now solved by an abstract state machine that gradually builds up the dependency tree token by token. 
The goal of this kind of parsers is to find the most probable transition sequence from an initial to some terminal configuration (\ie a dependency parse tree, or in our case a \emph{property tree}) given a permissible set of actions (\ie LEFT-ARC, RIGHT-ARC, SHIFT) and they are able to handle both projective and non-projective dependencies~\citep{nivre:03,nivre:09}. In the simplest case (\ie greedy inference), a classifier predicts the next transition based on the current configuration.
Compared to graph-based dependency parsers, transition-based parsers are able to scale better due to the linear time complexity while graph-based complexity rises to $O(n^2)$ in the non-projective case.~\cite{chen:14} proposed a way of learning a neural network classifier for use in a greedy, transition-based dependency parser while using low-dimensional, dense word embeddings, without the need of manually extracting features. 
Globally normalized transition-based parsers \citep{andor:16} can be considered an extension of \cite{chen:14}, as they perform beam search for maintaining multiple hypotheses and introduce global normalization with a CRF objective.~\cite{dyer:15} introduced the stack-LSTM model with push and pop operations which is able to learn the parser transition states while maintaining a summary embedding of its contents. 
Although transition-based systems are well-known for their speed and state-of-the-art performance, we do not include them in our study due to their already reported poor performance in the real estate task \citep{bekoulis:17} 
compared to graph-based parsers.

\subsection{Joint learning}
\label{sec:related_joint_learning}

\noindent 
Adopting a pipeline strategy for the considered type of problems has two main drawbacks:
\begin{enumerate*}[label=(\roman*)]
\item sequence labeling errors propagate to the dependency parsing step, \eg an incorrectly identified part of the house (entity) could get connected to a truly existing entity, and
\item interactions between the components are not taken into account (feedback between the subtasks), \eg modeling the relation between two potential entities may help in deciding on the nature of the entities themselves.
\end{enumerate*}
In more general relation extraction settings, a substantial amount of work~\citep{kate:10,li:14,miwa:14} jointly considered the two subtasks of entity recognition and relation extraction. However, all of these models make use of hand-crafted features that:
\begin{enumerate*}[label=(\roman*)]
\item require manual feature engineering,
\item generalize poorly between various applications and
\item may require a substantial computational cost
\end{enumerate*}.

Recent advances on joint models for general relation extraction consider the joint task using neural network architectures like LSTMs and CNNs~\citep{miwa:16,zheng:17,li:17}. Our work is however different from a typical relation extraction setup in that we aim to model directed spanning trees, 
or, equivalently, non-projective dependency structures. In particular, the entities involved in a relation are not necessarily adjacent in the text since other entities may be mentioned in between, which complicates parsing.
Indeed, in this work we focus on dependency parsing due to the difficulty of establishing the tree-like structure instead of only relation extraction (where each entity can have arbitrary relation arcs, regardless of other entities and their relations), which is 
the case for previously cited joint models.
Moreover, unlike most of these works that frame the problem as a stacking of the two components, or at least first train the NER module to recognize the entities and then further train together with the relation classification module, we include the NER directly inside the dependency parsing component.

In summary, the conceptual strengths of our \emph{joint} segmentation and dependency parsing approach (described in detail in \secref{sec:methodology}) will be the following: compared to state-of-the-art joint models in relation extraction, it 
\begin{enumerate*}[label=(\roman*)]
\item is generic in nature, without requiring any manual feature engineering, 
\item extracts a complete tree structure rather than a single binary relation instance.
\end{enumerate*}

\begin{table}
\resizebox{\columnwidth}{!}{%

\begin{tabular}{p{2.2cm}p{4.5cm}p{3.6cm}}
\toprule
Entity type   & Description  & Examples  \\
\midrule
$\mathsf{property}$ & The property. & bungalow, apartment\\
$\mathsf{floor}$ & A floor in a building. &  ground floor\\
$\mathsf{space}$ & A room within the building. & bedroom, bathroom\\
$\mathsf{subspace}$ & A part of a room. & shower, toilet \\
$\mathsf{field}$ & An open space inside or outside the building. & bbq, garden \\
$\mathsf{extra\ building}$ & An additional building which is also part of the property. & garden house\\
\bottomrule
\end{tabular}
}
\caption{Real estate entity types.}
\label{tab:entitytypes}
\end{table}

\section{Problem definition}
\label{problem_definition}
\noindent In this section, we define the specific terms that are used in our real estate structured prediction problem. We define an entity as an unambiguous, unique part of a property with independent existence (e.g., bedroom, kitchen, attic).
An entity mention is defined as one or more sequential tokens (\eg ``large apartment'') that can be potentially linked to one or more entities. An entity mention has a unique semantic meaning and refers to a specific entity, or a set of similar entities (\eg ``six bedrooms''). An entity itself is \emph{part-of} another entity and can be mentioned in the text more than once with different entity mentions. For instance, a ``house'' entity could occur in the text with entity mentions ``large villa'' and ``a newly built house''.
For the pipeline setting as presented in \cite{bekoulis:17}, we further classify entities into types (assign a
named entity type to every word in the ad). The task is transformed to a sequence labeling problem using BIO (Beginning, Inside, Outside) encoding. 
The entity types are listed in \tabref{tab:entitytypes}.
For instance, in the sequence of tokens ``large apartment'', B-PROPERTY is assigned to the token ``large'' as the beginning of the entity, I-PROPERTY in the token ``apartment'' as the inside of the entity but not the first token within the entity and O for all the other tokens that are not entities.
Unlike previous studies~\citep{miwa:16,zheng:17,li:16,li:17}, for our joint model there is no need for this type of categorical classification into labels since the two components are treated unified as a single dependency parsing problem.

\lstset{emph={Original ad:},emphstyle=\textbf}
\lstset{emph={Structured representation:},emphstyle=\textbf}

\begin{figure}
\begin{center}
\begin{tabular}{c}
\begin{adjustbox}{width=\textwidth,height=3.5cm,keepaspectratio}
\begin{lstlisting}[escapeinside={(*}{*)},style=base]
[*Original ad:*]
The (property( includes a @large apartment@ with a @garage@. The 
@home@ has a !living room!, !3 spacious bedrooms! and a !bathroom!.
The @garage@ is equipped with a !gate! and a !bike wall bracket!.
------------------------------------------
[*Structured representation:*]
(property			| mention=`property'(
   @large apartment		| mention=`large apartment', `home'@
      !living room		| mention=`living room'!
      !3 spacious bedrooms	| mention=`3 spacious bedrooms'!
      !bathroom			| mention=`bathroom'!
   @garage			| mention=`garage'@
      !gate			| mention=`gate'!
      !bike wall bracket		| mention=`bike wall bracket'!
    
\end{lstlisting}
\end{adjustbox}
\end{tabular}
\end{center}
\caption{Fictitious sample unstructured ad and corresponding structured representation as a property tree.}
\label{fig:plain_ad}
\end{figure}

The goal of the real estate structured prediction task is to map the textual property classified into a tree-like structured representation, the so-called \emph{property tree}, as illustrated in \figref{fig:plain_ad}. In the pipeline setting, this conversion implies the detection of
\begin{enumerate*}[label=(\roman*)]
\item entities of various types and
\item the \emph{part-of} dependencies between them.
\end{enumerate*}
For instance, the entity ``living room'' is \emph{part-of} the entity ``large apartment''.
In the joint model, each token (\eg ``apartment'', ``living'', ``bathroom'', ``includes'', ``with'', ``3'') is examined separately and 4 different types of relations are defined, namely \emph{part-of}, \emph{segment}, \emph{skip} and \emph{equivalent}.
The \emph{part-of} relation is similar to the way that it was defined in the pipeline setting but instead of examining entities, \ie sequences of tokens (\eg ``living room''), we examine if a (individual) token is \emph{part-of} another (individual) token (\eg ``room'' is \emph{part-of} the ``apartment'').
We encode the entity identification task with the \emph{segment} label and we follow the same approach as in the \emph{part-of} relationships for the joint model. Specifically, we examine if a token is a \emph{segment} of another token (\eg the token ``room'' is attached as a \emph{segment} to the token ``living'', ``3'' is attached as a \emph{segment} to the token ``bedrooms'' and ``spacious'' is also attached as a \emph{segment} to the token ``bedrooms'' --- this way we are able to encode the \emph{segment} ``3 spacious bedrooms''). By doing so, we cast the sequence labeling subtask to a dependency parsing problem. The tokens that are referring to the same entity belong to the \emph{equivalent} relation (``home'' is \emph{equivalent} to ``apartment''). For each entity, we define the first mention in order of appearance in the text as main mention and the rest as \emph{equivalent} to this main mention.
Finally, each token that does not have any of the aforementioned types of relations has a \emph{skip} relation with itself (\eg ``includes'' has a \emph{skip} relation with ``includes''), such that each token has a uniquely defined head. 

\begin{figure}
\centering				
				\includegraphics[width=0.60\textwidth,keepaspectratio]{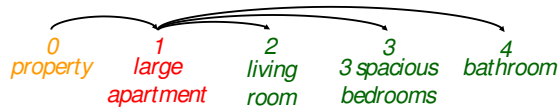}
				\caption{An example graph of projective \emph{part-of} dependencies.}
				\label{fig:projective}
			\end{figure}

\begin{figure}
				\centering
				\includegraphics[width=0.85\textwidth,keepaspectratio]{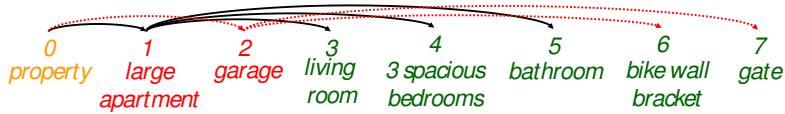}
				\captionsetup{justification=centering}

				\caption{Graph representing the \emph{part-of} dependencies of \figref{fig:plain_ad}. The dashed arcs are representing the non-projective dependencies.}
				\label{fig:non_projective}
			\end{figure}

Thus, we cast the structured prediction task of extracting the \emph{property tree} from the ad as a dependency parsing problem,
where
\begin{enumerate*}[label=(\roman*)]
\item an entity can be \emph{part-of} only one (other) entity, because the decisions are taken simultaneously for all \emph{part-of} relations (\eg a certain room can only be \emph{part-of} a single floor), and
\item there are a priori no restrictions on the type of entities or tokens that can be \emph{part-of} others (\eg a room can be either \emph{part-of} a floor, or the property itself, like an apartment).
\end{enumerate*}
It is worth mentioning that dependency annotations for our problem exhibit a significant number of 
non-projective arcs (26\%) where \emph{part-of} dependencies are allowed to cross (see \figref{fig:non_projective}), meaning that entities involved in the \emph{part-of} relation are non-adjacent (\ie interleaved by other entities). For instance, all the entities or the tokens for the pipeline and the joint models, that are attached to the entity ``garage'' are overlapping with the entities that are attached to the entity ``apartment'', making parsing even more complicated: handling only projective dependencies as illustrated in \figref{fig:projective} is an easier task. 
We note that the \emph{segment} dependencies do not suffer from non-projectivity, since the tokens are always adjacent and sequential (\eg ``3 spacious bedrooms'').

\section{Methodology}

\label{sec:methodology}
\noindent We now describe the two approaches, \ie the pipeline model and the joint model to construct the \emph{property tree} of the textual advertisements, as illustrated in \figref{fig:system_setup}. For the pipeline system (\secref{subsect:two-step_pipeline}), we
\begin{enumerate*}[label=(\arabic*)]
\item \label{step:sequence-labeling} identify the entity mentions (\secref{subsect:sequence_labeling}),
then
\item \label{step:relation-extraction} predict the \emph{part-of} dependencies between them (\secref{subsect:relation_extraction}), 
and finally
\item \label{step:tree-construction} construct the tree representation (\ie\emph{property tree}) of the textual classified (\eg as in \figref{fig:plain_ad}).
\end{enumerate*}
In step~\ref{step:relation-extraction}, we apply locally or globally trained graph-based models.  
We represent the result of step~\ref{step:relation-extraction} as a graph model, and then solve step~\ref{step:tree-construction}
by applying the maximum spanning tree algorithm~\citep{chu:65,edmond:68} for the directed case (see~\cite{mcdonald:05}).
We do not apply the well-known and fast transition-based systems with hand-crafted features for non-projective dependency structures \citep{nivre:09,bohnet:12}, given the previously established poor performance thereof in \cite{bekoulis:17}.
In \secref{subsect:joint_learning}, we describe the joint model where we perform steps~\ref{step:sequence-labeling} and \ref{step:relation-extraction} jointly. For step~\ref{step:tree-construction}, we apply the maximum spanning tree algorithm~\citep{chu:65,edmond:68} similarly as in the pipeline setting (\secref{subsect:two-step_pipeline}). 

\subsection{Two-step pipeline}
\label{subsect:two-step_pipeline}
\noindent Below we revisit the pipeline approach presented in \cite{bekoulis:17}, which serves as the baseline which we compare the neural models against.
As mentioned before, the pipeline model comprises two subtasks:
\ref{step:sequence-labeling}~the sequence labeling and the
\ref{step:relation-extraction}~\emph{part-of} tree construction. In the following subsections, we describe the methods applied for both. 

\subsubsection{Sequence labeling}
\label{subsect:sequence_labeling}

\noindent The first step in our pipeline approach is the sequence labeling subtask which is similar to NER. Assuming a textual real estate classified, we
\begin{enumerate*}[label=(\roman*)]
\item identify the entity mention boundaries and 
\item map each identified entity mention to a categorical label, \ie entity type.
\end{enumerate*}
In general, in the sequence labeling tasks, it is beneficial to take into account the correlations between labels in adjacent tokens, \ie consider the neighborhood, and jointly find the most probable chain of labels for the given input sentence (Viterbi algorithm for the most probable assignment). For instance, in our problem where we follow the NER standard BIO encoding~\citep{ratinov:09}, the I-PROPERTY cannot be followed by I-SPACE without first opening the type by B-SPACE.
We use a special case of the CRF algorithm~\citep{crf:01,peng:06}, namely linear chain CRFs, which is commonly applied in the problem of sequence labeling to learn a direct mapping from the feature space to
the output space (types) where we model label sequences jointly, instead of decoding each label independently. A linear-chain CRF
with parameters $w$ defines a conditional probability $P_w(y|x)$ for the sequence of labels $y = y_1,...,y_N$ given the tokens of the text advertisement $x =x_1,...,x_N$ to be 
\begin{equation}
P_w(y|x)=\frac{1}{Z(x)}\exp({w^T\phi(x,y)}),
\end{equation}
where $Z$ is the normalization constant and $\phi$ is the feature function that computes a feature vector given the advertisement and the sequence of labels.

\begin{figure}
\centering
\includegraphics[width=0.7\columnwidth]{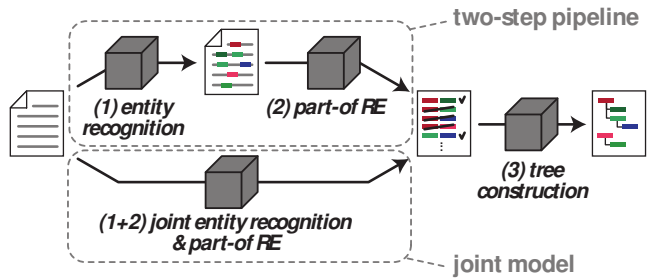}
\caption{The full structured prediction system setup.}
\label{fig:system_setup}
\end{figure}

\subsubsection{Part-of tree construction}
\label{subsect:relation_extraction}

\noindent The aim of the \emph{part-of} tree construction subtask is to link each entity to its parent. We approach the task as a dependency parsing problem but instead of connecting each token to its syntactical parent, we map only the entity set $I$ (\eg ``large villa'', ``3 spacious bedrooms'') that has already been extracted by the sequence labeling subtask to a dependency structure $y$. 
Assuming the entity set $I=\{e_0,e_1,...,e_t\}$ where $t$ is the number of identified entities, a dependency is a pair $(p,c)$ where $p\in I$ is the parent entity and $c\in I$ is the child entity. The entity $e_0$ is the dummy root-symbol that only appears as parent.


We will compare two approaches to predict the \emph{part-of} relations: a locally trained model (LTM) scoring all candidate edges independently, versus a global model (MTT) which jointly scores all edges as a whole.

\subsubsection*{Locally trained model (LTM)}

\noindent In the locally trained model (LTM), 
we adopt a traditional local discriminative method and apply a binary classification framework~\citep{yamada:03} to learn the \emph{part-of} relation model (step~\ref{step:relation-extraction}), based on standard relation extraction features such as the parent and child tokens and their types, the tokens in between, etc.
For each candidate parent-child pair, the classifier gives a score that indicates whether it is probable for the \emph{part-of} relation to hold between them. The output scores are then used for step~\ref{step:tree-construction}, to construct the final \emph{property tree}.
Following~\cite{mcdonald:05,mcdonald:06}, we view the entity set $I$ as a 
fully connected directed graph $G=\{V,E\}$ with the entities $e_1,..., e_t$ as vertices ($V$) in the graph $G$, and edges $E$ representing the \emph{part-of} relations with the respective classifier scores as weights. One way to approach the problem is the greedy inference method where the predictions are made independently for each parent-child pair, thus neglecting that the global target output should form a \emph{property tree}. We could adopt a threshold-based approach, \ie keep all edges exceeding a threshold, which obviously is not guaranteed to end up with arc dependencies that form a tree structure (\ie could even contain cycles).
On the other hand, we can enforce the tree structure inside the (directed) graph by finding the maximum spanning tree. To this end, similar to~\cite{mcdonald:05,mcdonald:06}, we apply the Edmonds' algorithm to search for the most probable non-projective tree structure in the weighted fully connected graph $G$.

\subsubsection*{Globally trained model (MTT)}
\noindent The Matrix-Tree theorem (MTT)~\citep{koo:07} is a globally normalized statistical method that involves the learning of directed spanning trees. Unlike the locally trained models, MTT is able to learn tree dependency structures, \ie scoring parse trees for a given sentence. We use $D(I)$ to refer to all possible dependencies of the identified entity set $I$, in which each dependency is represented as a tuple $(h,m)$ in which $h$ is the head (or parent) and $m$ the modifier (or child). The set of all possible dependency structures for a given entity set $I$ is written $T(I)$.
The conditional distribution over all dependency structures $y\in T(I)$ can then be defined as: 
\begin{equation} 
P(y | I;\theta ) = \frac{1}{Z(I;\theta)}\exp\left(\sum\limits_{h,m \in y} \theta_{h,m}\right) 
\end{equation} 
in which the coefficients $\theta_{h,m}\in\mathbb{R}$ for each dependency $(h,m)$ form the real-valued weight vector $\theta$.
%
The partition function $Z(I;\theta)$ 
is a normalization factor
that alas cannot be computed by brute-force, since it requires a summation over 
all $y\in T(I)$, containing
an exponential number of possible dependency structures. However, an adaptation of the MTT allows us the direct and efficient computation of the partition function $Z(I;\theta)$ as the determinant $det(L(\theta))$  where $L(\theta)$ is the Laplacian matrix of the graph. It is worth mentioning that although  MTT learns spanning tree structures during training, at the prediction phase, it is still required to use the maximum spanning tree algorithm (step~\ref{step:tree-construction})~\citep{mcdonald:05,mcdonald:06} as in the locally trained models.

\subsection{Joint model}
\label{subsect:joint_learning}
\begin{figure}
\centering
\includegraphics[width=\textwidth,keepaspectratio]{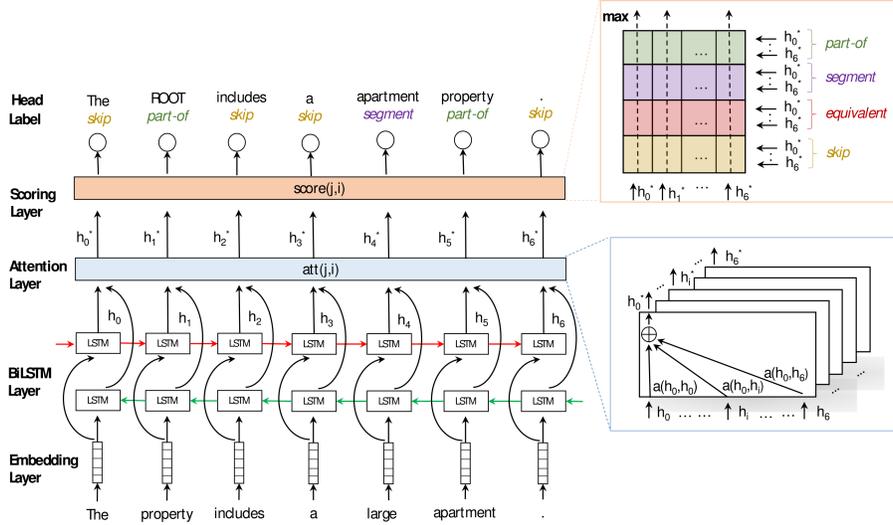}
\caption{The architecture of the joint model.}
\label{fig:joint_model}
\end{figure}

\noindent
In this section, we present the new joint model sketched in \figref{fig:joint_model}, which simultaneously predicts the entities in the sentence and the dependencies between them, 
with the final goal of obtaining a tree structure, i.e., the \emph{property tree}.
We pose the problem of the identification of the entity mentions and the dependency arcs between them as a head selection problem~\citep{zhang:16}. Specifically, given as input a sentence of length $N$, the model outputs the predicted parent of each token of the advertisement and the most likely dependency label between them. 
We begin by describing how the tokens are represented in the model, \ie with fixed pre-trained embeddings (\secref{subsec:embeddings}), which form the input to an LSTM layer (\secref{subsec:lstms}). The LSTM outputs are used as input to the entity and dependency scoring layer (\secref{subsec:head_selection}).
%
As an extension of this model, we propose the use of various attention layers in between the LSTM and scoring layer, to encourage the model to focus on salient information, as described in \secref{subsec:attention}. The final output of the joint model still is not guaranteed to form a tree structure. Therefore, we still apply Edmonds' algorithm (i.e., step~\ref{step:tree-construction} from the pipeline approach), described in \secref{subsec:edmond}. 

\subsubsection{Embedding Layer}
\label{subsec:embeddings}
\noindent
The embedding layer maps each
token of the input sequence ${x_1,...,x_N }$ of the considered advertisement to a low-dimensional vector space. We obtain the word-level embeddings by training the Skip-Gram word2vec model~\citep{mikolov:13} on a large collection of property advertisements.
We add a symbol $x_0$ in front of the $N$-length input sequence, which will act as the root of the property tree, and is represented with an all-zeros vector in the embedding layer. 

\subsubsection{Bidirectional LSTM encoding layer} 
\label{subsec:lstms}

\noindent
Many neural network architectures have been proposed in literature: LSTMs~\citep{hochreiter:97}, CNNs~\citep{lecun:89}, Echo State Networks~\citep{jaeger:01}, or 
Stochastic Configuration Networks~\citep{wang:17}, to name only a few. Many others can be found in reference works on the topic~\citep{goodfellow:16,goldberg:17}. 
In this work, we use RNNs which have been proven to be particularly effective in a number of NLP tasks~\citep{sutskever:14,lample:16,miwa:16}.
Indeed, RNNs are a common and reasonable choice to model sequential data and inherently able to cope with varying sequence lengths.
Yet, plain vanilla RNNs tend to suffer from vanishing/exploding gradient problems and are hence not successful in capturing long-term dependencies~\citep{bengio:94,pascanu:13}.
LSTMs are a more advanced kind of RNNs, which have been successfully applied in several tasks to capture long-term dependencies, as they are able to effectively overcome the vanishing gradient problem.
For many NLP tasks, it is crucial to represent each word in its own context, \ie to consider both past (left) and future (right) neighboring information. An effective solution to achieve this is using a bidirectional LSTM (BiLSTM). The basic idea is to encode each sequence from left to right (forward) and from right to left (backward). This way, there is one hidden state which represents the past information and another one for the future information. The high-level formulation of an LSTM is:
\begin{equation}
h_i,c_i= \text{LSTM}(w_i,h_{i-1},c_{i-1}), \;\;i=0,...,N
\end{equation}
where in our setup $w_i \in \mathbb{R}^{\tilde{d}}$ is the word embedding for token $x_i$, and with the  input and states for the root symbol $x_0$ initialized as zero vectors. Further, $h_i \in \mathbb{R}^d$ and $c_i \in \mathbb{R}^d$ respectively are the output and cell state for the $i$th position, where $d$ is the hidden state size of the LSTM. Note that we chose the word embedding size the same as the LSTM hidden state size, or $\tilde{d}=d$. The outputs from left to right (forward) are written as $\vec{h_i}$ and the outputs from the backwards direction as $\cev{h_i}$.
The two LSTMs' outputs at position $i$ are concatenated to form the output $h_i$ at that position of the BiLSTM:
\begin{equation}
h_i= [\vec{h_i};\cev{h_i}],\;\; i=0,...,N
\end{equation}

\subsubsection{Joint learning as head selection}
\label{subsec:head_selection}

\noindent In this subsection, we describe the joint learning task (\ie identifying entities and predicting dependencies between them), which we formulate as a head selection problem~\citep{zhang:16}. Indeed, each word $x_i$ should have a unique head (parent) --- while it can have multiple dependent words --- since the final output should form the \emph{property tree}. Unlike the standard head selection dependency parsing framework \citep{zhang:16}, we predict the head $y_i$ of each word $x_i$ and the relation $c_i$ between them 
jointly, instead of first obtaining binary predictions for unlabeled dependencies, followed by an additional classifier to predict the labels. 

Given a text advertisement as a token sequence $x = x_0, x_1,..., x_N$ where $x_0$ is the dummy root symbol, and a set $\mathcal{C} = \{\emph{part-of}, \emph{segment}, \emph{equivalent}, \emph{skip}\}$  of predefined labels (as defined in \secref{problem_definition}), we aim to find for
each token $x_i,\;i\in\{0,...,N\}$  
the most probable head $x_j,\;j\in\{0,...,N\}$  
and the most probable corresponding label $c\in \mathcal{C}$.  For convenience, we order the labels $c\in\mathcal{C}$ and identify them as $c_k,\;k\in\{0,...,3\}$. We model the joint probability of token $x_j$ to be the head of $x_i$ with $c_k$ the relation between them, using a softmax:  
\begin{equation}
P(head=x_j,label=c_k |x_i)=\frac{\exp(score(h_j , h_i,c_k))}{\sum_{\tilde{j},\tilde{k}}\exp(score(h_{\tilde{j}} , h_i,c_{\tilde{k}})}
\end{equation}
where $h_i$ and $h_j$ are the BiLSTM encodings for words $x_i$ and $x_j$, respectively. For the scoring formula $score(h_j , h_i,c_k)$ we use a neural network layer that computes the relative score between position $i$ and $j$ for a specific label $c_k$ as follows:
\begin{equation}
score(h_j , h_i,c_k)= V_k ^T \tanh (U_k  h_j + W_k  h_i+b_k)
\end{equation}
with trainable parameters $V_k \in \mathbb{R}^{l}$, $U_k \in \mathbb{R}^{l \times 2d}$, $W_k \in \mathbb{R}^{l \times 2d}$, $b_k \in \mathbb{R}^{l}$, and $l$ the layer width.
As detailed in \secref{sec:setup}, we set $l$ to be smaller than $2d$, similar to \cite{dozat:16} due to the fact that training on superfluous information reduces the parsing speed and increases tendency towards overfitting. 
We train our model by minimizing the cross-entropy loss $\mathcal{L}$, written for the considered training instance as:
\begin{equation}
\mathcal{L}=\sum_{i=0}^{N} -\log P (head=y_i,label=c_i|x_i)
\end{equation}
where $y_i\in x$ and $c_i\in\mathcal{C}$ are the ground truth head and label of $x_i$, respectively. After training, we follow a greedy inference approach and for each token, we simultaneously keep the highest scoring head $\hat{y_i}$ and label $\hat{c_i}$ for $x_i$ based on their estimated joint probability:
\begin{equation}
(\hat{y_i},\hat{c_i})=\argmax_{x_j \in x,c_k \in \mathcal{C}} P(head=x_j,label=c_k|x_i)
\end{equation}
The predictions $(\hat{y_i},\hat{c_i})$ are made independently for each position $i$, neglecting that the final structure should be a tree. Nonetheless, as demonstrated in \secref{sec:comparison_pipeline_joint}, the highest scoring neural models are still able to come up with a tree structure for 78\% of the ads. In order to ensure a tree output in all cases, however, we apply Edmonds' algorithm on the output.

\subsubsection{Attention Layer}
\label{subsec:attention}

\noindent The attention mechanism in our structured prediction problem aims to improve the
model performance by focusing on information that
is relevant to the prediction of the most probable head for each token. As attention vector, we construct the new context vector $h_i^*$ as a weighted average of the BiLSTM outputs  
\begin{equation}
h_j^* = \sum_{i=0}^{N} a(h_j,h_i) \> h_i
\end{equation}
in which the coefficients $a(h_j, h_i)$, also called the attention weights, are obtained as follows:
\begin{equation}
a(h_j,h_i)=\frac{\exp(att(h_j,h_i))}{\sum_{\tilde{i}=0}^{N} \exp(att(h_{j},h_{\tilde{i}}))}.
\end{equation}
The attention function $att(h_j, h_i)$ is designed to measure some form of compatibility between the representation $h_i$ for $x_i$ and $h_j$ for $x_j$, and the attention weights $a(h_j, h_i)$ are obtained from these scores by normalization using a softmax function.
%
In the following, we will describe in detail the various attention models that we tested with our joint model.
\subsubsection*{Commonly used attention mechanisms}
\noindent Three commonly used attention mechanisms are listed in \cref{eq:additive,eq:bilinear,eq:multiplicative}: the additive~\citep{vinyals:15}, bilinear, and multiplicative attention models~\citep{luong:15}, which have been extensively used in machine translation. Given the representations $h_i$ and $h_j$ for tokens $x_i$ and $x_j$, we compute the attention scores as follows:
\begin{equation}
att_\textit{additive}(h_j , h_i) = V_a \tanh (U_a  h_j + W_a  h_i+b_a)
\label{eq:additive}
\end{equation}
\begin{equation}
att_\textit{bilinear}(h_j , h_i) = h_j^T W_\textit{bil} h_i
\label{eq:bilinear}
\end{equation}
\begin{equation}
att_\textit{multiplicative}(h_j , h_i) = h_j^T  h_i
\label{eq:multiplicative}
\end{equation}
where $V_a \in \mathbb{R}^{l}$, $U_a, W_a \in \mathbb{R}^{l \times 2d}$, $W_{bil} \in \mathbb{R}^{2d \times 2d}$ and $b_a \in \mathbb{R}^{l}$ are learnable parameters of the model. 

\subsubsection*{Biaffine attention}
\noindent We use the biaffine attention model~\citep{dozat:16} which has been recently applied to dependency parsing and is a modification of the neural graph-based approach that was proposed by~\cite{kiperwasser:16}. In this model,~\cite{dozat:16} tried to reduce the dimensionality of the recurrent state of the LSTMs by applying a such neural network layer on top of them. This idea is based on the fact that there is redundant information in every hidden state that (i)~reduces parsing speed and (ii)~increases the risk of overfitting. To address these issues, they reduce the dimensionality and apply a nonlinearity afterwards. The deep bilinear attention mechanism is defined as follows:
\begin{equation}
h_i^\textit{dep}=V_\textit{dep} \tanh (U_\textit{dep}  h_i+b_\textit{dep})
\end{equation}
\begin{equation}
h_j^\textit{head}=V_\textit{head} \tanh (U_\textit{head}  h_j+b_\textit{head})
\end{equation}
\begin{equation}
att_\textit{biaffine}(h_j^\textit{head} , h_i^\textit{dep}) =(h_j^\textit{head})^T W_\textit{bil} h_i^\textit{dep} +Bh_j^\textit{head}
\end{equation}
where $U_\textit{dep}, U_\textit{head} \in \mathbb{R}^{l \times 2d}$, $V_\textit{dep}, V_\textit{head} \in \mathbb{R}^{p \times l}$, $W_\textit{bil} \in \mathbb{R}^{p \times p}$, $B \in \mathbb{R}^{p}$ and $b_\textit{dep}$, $b_\textit{head} \in \mathbb{R}^{l}$.
\subsubsection*{Tensor attention}
\noindent This section introduces the Neural Tensor Network~\citep{socher:13} that has been used as a scoring formula applied for relation classification between entities. The task can be described as link prediction between entities in an existing network of relationships. We apply the tensor scoring formula as if tokens are entities, by the following function:
\begin{equation}
att_\textit{tensor}(h_j,h_i)=U_t\tanh(h_j^TW_th_i+V_t( h_j + h_i)+b_t)
\end{equation}
where $W_t \in \mathbb{R}^{2d \times l \times 2d}$, $V_t \in \mathbb{R}^{l \times 2d}$, $U_t \in \mathbb{R}^{l}$ and $b_t\in \mathbb{R}^{l}$.

\subsubsection*{Edge attention}
\noindent In the edge attention model, we are inspired by \cite{gilmer:17}, which applies neural message passing in chemical structures. Assuming that words are nodes inside the graph and the message flows from node $x_i$ to $x_j$, we define the edge representation to be:
\begin{equation}
\textit{edge}(h_j,h_i)=\tanh (U_e  h_j + W_e  h_i+b_e)
\end{equation}
The edge attention formula is computed as:
\begin{equation}
att_\textit{edge}(h_j,h_i)=\frac{1}{N}\left(A_\textit{src} \sum_{\tilde{i}=0}^{N} \textit{edge}(h_j,h_{\tilde{i}}) +A_\textit{dst} \sum_{\tilde{j}=0}^{N} \textit{edge}(h_{\tilde{j}},h_i)\right)
\end{equation}
where $U_e, W_e \in \mathbb{R}^{l \times 2d}$, $A_\textit{src}, A_\textit{dst} \in \mathbb{R}^{2d \times l}$ and $b_e \in \mathbb{R}^{l}$. The source and destination matrices respectively encode information for the start to the end node, in the directed edge. Running the edge attention model for several times can be achieved by stacking the edge attention layer multiple times. This is known as message passing phase and we can run it for several ($T > 1$)  time steps to obtain more informative edge representations.

\subsubsection{Tree construction step: Edmonds' algorithm}
\label{subsec:edmond}

\noindent At decoding time, greedy inference is not guaranteed to end up with arc dependencies that form a tree structure and the classification decision might contain cycles. In this case, the output can be post-processed with a maximum spanning tree algorithm (as the third step in \figref{fig:system_setup}). We construct the fully connected directed graph $G = (V, E)$ where the vertices $V$ are the tokens of the advertisement (that are not predicted as \emph{skips}) and the dummy root symbol, $E$ contains the edges representing the highest scoring relation (e.g., \emph{part-of}, \emph{segment}, \emph{equivalent}) with the respective cross entropy scores serving as weights. Since $G$ is a directed graph, $s(x_i, x_j)$ is not necessarily equal to $s(x_j, x_i)$. Similar to \cite {mcdonald:05}, we employ Edmonds' maximum spanning tree algorithm for directed graphs \citep{chu:65,edmond:68} to build a non-projective parser. Indeed, in our setting, we have a significant number (26\% in the dataset used for experiments, see further) of non-adjacent \emph{part-of} and \emph{equivalent} relations (non-projective). It is worth noting that in the case of \emph{segment} relations, the words involved are not interleaved by other tokens and are always adjacent. We apply Edmonds' algorithm to every graph which is constructed to get the highest scoring graph structure, even in the cases where a tree is already formed by greedy inference. For \emph{skips}, we consider the predictions as obtained from the greedy approach and we do not include them in the fully connected weighted graph, since Edmonds' complexity is $O(n^2)$ for dense graphs and might lead to slow decoding time.

\section{Results and discussion}
\label{sec:results_discussion}
\noindent In this section, we present the experimental results of our study. We describe the dataset, the setup of the experiments and we compare the results of the methods analysed in the previous sections.

\subsection{Experimental setup}
\label{sec:setup}
\noindent Our dataset consists of a large collection (\ie 887,599) of Dutch property advertisements from real estate agency websites. From this large dataset, a sub-collection of 2,318 classifieds have been manually annotated by 3 trained human annotators (1 annotation per ad, 773 ads per annotator). The annotations follow the format of the \emph{property tree} that is described in detail in \secref{problem_definition} and is illustrated in \figref{fig:plain_ad}. The dataset is available for research purposes, see our github codebase.\footnote{\label{ft:github}\url{https://github.com/bekou/ad_data}} In the experiments, we use only the annotated text advertisements for the pipeline setting, \ie LTM (locally trained model), MTT (globally trained model). In the case of the neural network approach, we train the embeddings on the large collection by using the word2vec model~\citep{mikolov:13} whereas in the joint learning, we use only the annotated documents, similar to the pipeline approach. 
The code of the LTM and the MTT hand-crafted systems is available on github.\textsuperscript{\ref{ft:github}} 
We also use our own CRF implementation. The code for the joint model has been developed in Python with the Tensorflow machine learning library~\citep{abadi:16} and will be made public as well.
For the evaluation, we use 70\% for training, 15\% for validation and 15\% as test set. We measure the performance by computing the $F_1$ score on the test set. The accuracy metric can be misleading in our case since we have to deal with imbalanced data (the \emph{skip} label is over-represented). We only report numbers on the structured classes, \ie \emph{segment} and \emph{part-of} since the other dependencies (\emph{skip}, \emph{equivalent}) are auxiliary in the joint models and do not directly contribute to the construction of the actual \emph{property tree}. For the overall $F_1$, we are again only considering the structured classes. Finally, we report the number of \emph{property trees} (which shows how likely our model is to produce trees without applying Edmonds' algorithm, \ie by greedy inference alone) for all the models before applying Edmonds' algorithm that guarantees the tree structure of the predictions.

For the pipeline models, we train the CRF with regularization parameter $\lambda_{CRF}=10$ and the LTM and MTT with $C=1$ based on the best hyperparameters on the validation set. As binary classifier, we use logistic regression. For the joint model, we train 128-dimensional word2vec embeddings on a collection of 887k advertisements. In general, using larger embeddings dimensions (\eg 300), does not affect the performance of our models. We consistently used single-layer LSTMs through our experiments to keep our model relatively simple and to evaluate the various attention methods on top of that. We have also reported results on the joint model using a two-layer stacked LSTM joint model,  
although it needs a higher computation time compared to a single-layer LSTM with an attention layer on top.
The hidden size of the LSTMs is $d=128$ and the size of the neural network used in the scoring and the attention layer is fixed to $l=32$. The optimization algorithm used is Adam~\citep{kingma:14} with a learning rate of $10^{-3}$. To reduce the effect of overfitting, we regularize our model using the dropout method~\citep{srivastava:14}. We fix the dropout rate on the input of the LSTM layer to 0.5 to obtain significant improvements ($\sim$1\%-2\% $F_1$ score increase, depending on the model). For the two-layer LSTM, we fix the dropout rate to 0.3 in each of the input layers since this leads to largest performance increase on the validation set. We have also explored gradient clipping without further improvement on our results. In the joint model setting, we follow the evaluation strategy of early stopping~\citep{caruana:01,graves:13} based on the performance of the validation set. In most of the experiments, we obtain the best hyperparameters after $\sim$60 epochs. 


\begin{table}
\resizebox{\columnwidth}{!}{%

\begin{tabular}{@{\extracolsep{4pt}}cccccccccc@{}} 
 \toprule
& \multicolumn{1}{c}{} &  \multicolumn{2}{c}{Precision}&  \multicolumn{2}{c}{Recall} &  \multicolumn{3}{c}{$F_1$ (\%)}&  \multicolumn{1}{c}{Trees} \\
\cline{3-4}
\cline{5-6}
\cline{7-9}

 & \multicolumn{1}{c}{Model} & \multicolumn{1}{c}{segment} & \multicolumn{1}{c}{part-of}& \multicolumn{1}{c}{segment} & \multicolumn{1}{c}{part-of} & \multicolumn{1}{c}{segment} & \multicolumn{1}{c}{part-of} & \multicolumn{1}{c}{Overall}& \multicolumn{1}{c}{(\% of ads)} \\

\midrule
\parbox[c]{8mm}{\multirow{2}{*}{\rotatebox[origin=c]{90}{\parbox{1.05cm}{\centering Hand-\\crafted }}}}

&LTM &   73.77    & 60.53&   70.98    & 60.40   &  72.35    &   60.47      &    64.76&   37.18 \\
&MTT&   73.77    & 61.15&   70.98    & 61.01   &  72.35    &  61.08      &    65.15&    43.23 \\

\midrule
\parbox[c]{8mm}{\multirow{3}{*}{\rotatebox[origin=c]{90}{\parbox{1.45cm}{\centering LSTM}}}}

  &LSTM&   70.24    &65.23&  77.73    &70.32      &   73.80      &    67.68&  68.82 &    68.30  \\
  &LSTM+E &   70.18    & 63.92&   77.77    & 71.08   &   73.78    &  67.31      &    68.57 &    68.30 \\
  &2xLSTM+E&   73.91    & 69.88 &   75.78    & 71.22   &  74.83    &   \textbf{70.54}      &     \textbf{70.90} &    78.09 \\

\midrule
  \parbox[c]{8mm}{\multirow{8}{*}{\rotatebox[origin=c]{90}{\parbox{1.4cm}{\centering Attentive \\ LSTM}}}} 
 &Additive&   72.97    & 65.71&   76.45    & 70.90      &  74.67  &   68.21&   69.46 &    74.35  \\
 
&Bilinear&   70.25    & 66.34&   79.96    & 72.53      &   74.79      &    69.29&  70.20 &   72.62  \\

&Multiplicative&   71.12    & 66.40&   77.81    &  71.26      &   74.31     &    68.75 &  69.70 &    72.91  \\

&Biaffine&    70.01    &  64.67&  78.32    & 71.04      &   73.93     &    67.71&  68.75 &    74.06  \\
&Tensor&   71.53    & 64.68&   76.17    & 70.79      &   73.78      &    67.60&  68.68 &    69.16  \\
&Edge$_1$&    71.56    & 67.46&   78.24    & 71.31      &    74.75      &   69.33 &  70.08 &    70.32  \\
    
&Edge$_2$&   72.03    & 66.09&   75.35    & 70.99     & 73.65     &    68.46&  69.12 &    73.48  \\     
          
     &Edge$_3$&   71.74    & 67.69&   78.44    & 73.00      &  \textbf{74.94}      &    70.25&  70.70 &    \textbf{78.96}  \\
     
\bottomrule
\end{tabular}
 }
 \caption{Performance of the three approaches on the structured prediction task. The top rows are for the pipeline approach, \ie \emph{hand-crafted} features. The next block of results presents the results for the neural joint model based on LSTMs. The bottom block contains results of the joint models augmented with several attentive architectures. Edmonds' algorithm is applied in all of the models to retain the tree structure, except for the LSTM joint model. The LSTM+E is the LSTM model with Edmonds' algorithm included. The 2xLSTM+E is the same joint model but it simply uses a stack of two LSTM layers. In the experiments with attention, we use a one-stack LSTM. The rightmost column is the percentage of the ads that are valid trees before applying Edmonds' (\ie step (3) of \figref{fig:system_setup}), showing the ability of the model to form trees during greedy inference. In the Edge$_i$ models, the number $i$ stands for the number of times that we have run the message passing phase.}
\label{tab:part_of_segments}
 \end{table}

\subsection{Comparison of the pipeline and the joint model}\label{sec:comparison_pipeline_joint}
\noindent One of the main contributions of our study is the comparison of the pipeline approach and the proposed joint model. We formulated the problem of identifying the entities (\ie \emph{segments}) and predicting the dependencies between them (\ie construction of the \emph{property tree}) as a joint model. Our neural model, unlike recent studies~\citep{miwa:16,zheng:17} on joint models that use LSTMs to handle similar tasks, does not need two components to model the problem (\ie NER and dependency parsing). To the best of our knowledge, our study is the first that formulates the task in an actual joint setting without the need to pre-train the sequence labeling component or for parameter sharing between them, since we use only one component for both subtasks. In \Tabref{tab:part_of_segments}, we present the results of the pipeline model (hand-crafted) and the proposed joint model (LSTM). The improvement of the joint model over the pipeline is unambiguous, \ie 3.42\% overall $F_1$ score difference between MTT (highest scoring pipeline model) and LSTM+E (LSTM model with Edmonds' algorithm). An additional increase of $\sim$2.3\% is achieved when we consider two-layer LSTMs (2xLSTM+E) for our joint model. All results in \Tabref{tab:part_of_segments}, except for the LSTM, are presented using Edmonds' algorithm on top, to construct the \emph{property tree}. Examining each label separately, we observe that the original LSTM+E model (73.78\%) performs better by 1.43\% in entity segmentation than the CRF (72.35\%). The LSTM model achieves better performance in the entity recognition task since it has to learn the two subtasks simultaneously resulting in interactions between the components (\ie NER and dependency parser). This way, the decisions for the entity recognition can benefit from predictions that are made for the \emph{part-of} relations. Concerning the \emph{part-of} dependencies, we note that the LSTMs outperform the hand-crafted approaches by 6.23\%. Also, the number of valid trees that are constructed before applying Edmonds' algorithm is almost twice as high for the LSTM models. Stacking two-layer LSTMs results in an additional $\sim$1\% improvement in the segmentation task and $\sim$3\% in the \emph{part-of} relations. The greedy inference for the hand-crafted methods does not produce well-formed trees, meaning that post-processing with Edmonds' algorithm (enforce tree structure) is expected to increase the performance of the hand-crafted models compared to the LSTM model performance. Indeed, the performance of the feature based hand-crafted models (\ie LTM and MTT) without the Edmonds' on top is not reported in \Tabref{tab:part_of_segments} due to their poor performance in our task (\ie $\sim$60\% overall $F_1$ and $\sim$51\% for \emph{part-of}), but after post-processing with Edmonds' the performance significantly increases (\ie$\sim$65\%). On the other hand, applying the Edmonds' algorithm on the LSTM model leads to marginally decreased performance ($\sim$0.2\%) compared to the original LSTM model, probably indicating that enforcing structural constraints is not beneficial for a model that clearly has the ability to form valid tree structures during greedy inference. Although one might be tempted not to enforce the tree structure (post-process with Edmonds'), due to the nature of our problem, we have to enforce tree constraints in all of the models.

\subsection{Comparison of the joint and the attention model}
\noindent After having established the superior performance of neural approach using LSTMs over the more traditional (LTM and MTT) methods based on hand-crafted features, we now discuss further improvements using attentive models. The attention mechanisms are designed to encourage the joint model to focus on informative tokens. We exploited several attention mechanisms as presented in \secref{subsec:attention}. \Tabref{tab:part_of_segments} shows the performance of the various models. Overall, the attention models are performing better in terms of overall $F_1$ score compared to the original joint model with the Edmonds' on top. Although the performance of the Biaffine and the Tensor models is limited compared to the improvement of the other attentive models, we focus on: 
\begin{enumerate*}[label=(\roman*)]
\item the Biaffine model since it achieved state-of-the-art performance on the dependency parsing task and 
\item the Tensor model because we were expecting that it would perform similarly to the Bilinear model (it has a bilinear tensor layer). 
\end{enumerate*} 
Despite its simplicity, the Bilinear model is the second best performing attentive model in \Tabref{tab:part_of_segments} in terms of overall $F_1$ score.
Edge$_3$ (70.70\% overall $F_1$ score) achieves better results than the other attention mechanisms in the entity recognition and in the dependency parsing tasks. We observe that running the message passing step  multiple times in the Edge model, gives an increasing trend in the number of valid trees that were constructed before applying the maximum spanning tree algorithm. This is not surprising since we expect that running the message passing phase multiple times leads to 
improved edge representations.
The maximum number of trees without post-processing by Edmonds' is attained when we run the message passing for 3 times whereas further increasing the number beyond 3 (\eg 4) appears no longer beneficial. Stacking a second LSTM layer on top of the joint model (2xLSTM+E) marginally improves the performance by 0.2\% compared to the Edge$_3$ attention model. But adding a second LSTM layer comes with the additional cost of an increased computation time compared to the joint models with the attention layers on top. This illustrates that:
\begin{enumerate*}[label=(\roman*)]
\item there might be some room for marginally improving the attention models even further, and
\item we do not have to worry about the quadratic nature of our approach since in terms of speed the attentive models are able to surpass the two-layer LSTMs.
\end{enumerate*} The sequential processing of the LSTMs might be the reason that slows down the computation time for the 2xLSTM over the rest of the attentive models. Specifically, on an Intel(R) Xeon(R) CPU E5-2650 v2 @ 2.60GHz processor, the best performing model (\ie Edge$_3$) takes $\sim$2 minutes per epoch while in the 2xLSTM case, it takes $\sim$2.5 minutes leading to a slowdown of $\sim$25\%. The percentage of the ads that are valid trees is 1\% better in the Edge$_3$ over the two-layer LSTM showcasing the ability of the Edge model to form more valid trees during greedy inference.

\subsection{Discussion}
\label{sec:discussion}
\noindent In this section, we discuss some additional aspects of our problem and the approaches that we follow
to handle them. 
As we mentioned before, a single entity can be present in the text with multiple mentions. 
This brings an extra difficulty to our task. 
For instance, in the example of \figref{fig:plain_ad}, the entity ``large apartment'' is expressed in the ad with the mentions ``large apartment'' and ``home''. 
Hence it is confusing to which mention the other entities should be attached to. One way would be to attach them to both and then eliminate one of the connections using Edmonds' spanning tree algorithm, which is the approach adopted in \cite{bekoulis:17}. 
The problematic issue with this approach is that the spanning tree algorithm would randomly remove all mentions but one, possibly resulting in uncertain outcomes. 
To avoid this problem, we now use as the main mention for an entity the first mention in order of appearance in the 
text (\eg ``large apartment'' in our example) and the remaining mentions (\eg ``home'') are attached as \emph{equivalent}
mentions to the main one. 
Usually, the most informative mention for an entity is the one that appears first, because we again refer to an entity mentioned before, often with a shorter description. 
Following our intuition, the neural model increases its overall performance by $\sim$3\% (from 66\% to 69\% and more than 5\% in the \emph{part-of} relation) and the pipeline approaches by almost 4\% (from 61\%, reported in \cite{bekoulis:17} to 65\% and more than 5\% in the \emph{part-of} relation). 

We also experimented with introducing the \emph{equivalent} relations. Although it is a strongly under-represented class in the dataset and the model performs poorly for this label (an \emph{equivalent} edge $F_1$ score of 10\%), introducing the \emph{equivalent} label is the natural way of modeling our problem (\ie assigning each additional mention as \emph{equivalent} to the main mention). We find out that introducing this type of relation leads to a slight decrease ($\sim$1\%) in the \emph{part-of} and a marginal increase ($\sim$0.3\%) in the \emph{segment} relations which are the main relations while retaining the nature of our problem. In the pipeline approach, it results in an 9\% drop in the $F_1$ score of the \emph{part-of} relation. This is the reason that the results as presented in \Tabref{tab:part_of_segments} do not consider the \emph{equivalent} relation for the hand-crafted model to make a fair comparison in the structured classes.

We believe our experimental comparison of the various architectural model variations provides useful findings for practitioners.
Specifically, for applications requiring both segmentation (entity recognition) and dependency parsing (structured prediction), our findings can be qualitatively summarized as follows:
\begin{enumerate*}[label=(\roman*)]
\item joint modeling is the most appropriate approach since it reduces error propagation between the components,
\item the LSTM model is much more effective (than models relying on handcrafted features) because it automatically extracts informative features from the raw text,
\item attentive models are proven effective because they encourage the model to focus on salient tokens,
\item the edge attention model leads to an improved performance since it better encodes the information flow between the entities by using graph representations, and
\item stacking a second LSTM marginally increases the performance, suggesting that there might be some room for slight improvement of the attention models by adding LSTM layers.
\end{enumerate*}

Finally, we point out how exactly our model relates to state-of-the-art in the field. Our joint model is able to both extract entity mentions (\ie perform segmentation) and do dependency parsing, which we demonstrate on the real estate problem. Previous studies~\citep{kate:10,li:14,miwa:14} that jointly considered the two subtasks (\ie segmentation and relation extraction):
\begin{enumerate*}[label=(\roman*)]
\item require manual feature engineering and
\item generalize poorly between various applications.
\end{enumerate*}
On the other hand, in our work, we rely on neural network methods (\ie LSTMs) to automatically extract features from the real estate textual descriptions and perform the two tasks jointly.
Although there are other methods which use neural network architectures~\citep{miwa:16,zheng:17,li:17} that focus on the relation extraction problem, our work is different in that we aim to model directed spanning trees and thus to solve the dependency parsing problem which is more constrained and difficult (than extracting single instances of binary relations). Moreover, the cited methods require either parameter sharing or pre-training of the segmentation module, which complicates learning. Therefore, cited methods are not directly comparable to our model and cannot be applied to our real estate task out-of-the-box.
However, our model's main limitation is the quadratic scoring layer that increases the time complexity of the segmentation task from linear (which is the complexity of a conditional random field, CRF) to $O(n^2)$. As a result, it sacrifices standard linear complexity of the segmentation task, in order to reduce the error propagation between the subtasks and thus perform learning in a joint, end-to-end differentiable, setting.

\section{Conclusions}
\label{sec:conclusions}
\noindent In this paper, we proposed an LSTM-based neural model to jointly perform \emph{segmentation} and \emph{dependency parsing}. We apply it to a real estate use case processing textual ads, thus \begin{enumerate*}[label=(\arabic*)]
\item identifying important entities of the property (\eg rooms) and
\item structuring them into a tree format
\end{enumerate*} 
based on the natural language description of the property. We compared our model with the traditional pipeline approaches that have been adapted to our task and we reported an improvement of 3.4\% overall edge $F_1$ score. Moreover, we experimented with different attentive architectures and stacking of a second LSTM layer over our basic joint model. The results indicate that exploiting attention mechanisms that encourage our model to focus on informative tokens, improves the model performance (increase of overall edge $F_1$ score with $\sim$2.1\%) and increases the ability to form valid trees in the prediction phase (4\% to 10\% more valid trees for the two best scoring attention mechanisms) before applying the maximum spanning tree algorithm. 

The contribution of this study to the research in expert and intelligent systems is three-fold:
\begin{enumerate*}[label=(\roman*)]
\item we introduce a generic joint model, simultaneously solving both subtasks of \emph{segmentation} (\ie entity extraction) and \emph{dependency parsing} (\ie extracting relationships among entities), that unlike previous work in the field does not rely on manually engineered features, 
\item in particular for the real estate domain, extracting a structured \emph{property tree} from a textual ad, we refine the annotations and additionally propose attention models, compared to initial work on this application, and finally
\item we demonstrate the effectiveness of our proposed generic joint model with extensive experiments (see aforementioned $F_1$ improvement of 2.1\%).
\end{enumerate*}
Despite the experimental focus on the real estate domain, we stress that the model is generic in nature, and could be equally applied to other expert system scenarios requiring the general tasks of both
 detecting entities (\emph{segmentation}) and
 establishing relations among them (\emph{dependency parsing}).
We furthermore note that our model, rather than focusing on extracting a single binary relation from a sentence (as in traditional relation extraction settings), produces a complete tree structure.

Future work can evaluate the value of our joint model we introduced in other specific application domains (\eg biology, medicine, news) for expert and intelligent systems.
For example, the method can be evaluated for entity recognition and binary relation extraction (the ACE 04 and ACE 05 datasets; see \citet{miwa:16}) or in adverse drug effects from biomedical texts (see \citet{li:16}).
In terms of model extensions and improvements, one research issue is to address the time complexity of the NER part by modifying the quadratic scoring layer for this component.
An additional research direction is to investigate different loss functions for the NER component (\eg adopting a conditional random field (CRF) approach), since this has been proven effective in the NER task on its own~\citep{lample:16}.
A final extension we envision is to enable multi-label classification of relations among entity pairs.

\section*{Acknowledgments}

\noindent The presented research was partly performed within the MALIBU project, funded by Flanders Innovation \& Entrepreneurship (VLAIO) contract number IWT 150630.


\bibliography{mybibfile}

\end{document}